\documentclass{article}

\usepackage[preprint,nonatbib]{neurips_2020}

\usepackage{algorithm}
\usepackage[noend]{algpseudocode}
\usepackage{times}
\usepackage{soul}
\usepackage{url}
\usepackage[hidelinks]{hyperref}
\usepackage[utf8]{inputenc}
\usepackage[T1]{fontenc}    % use 8-bit T1 fonts
\usepackage[small]{caption}
\usepackage{graphicx}
\usepackage{amsmath}
\usepackage{booktabs}
\usepackage{graphicx}
\usepackage{subfigure}
\usepackage{booktabs} % for professional tables
\usepackage{times}
\usepackage{epsfig}
\usepackage{graphicx}
\usepackage{amsmath}
\usepackage{amssymb}
\usepackage{xcolor}
\usepackage{soul}
\usepackage{booktabs}       % professional-quality tables
\usepackage{caption}
\usepackage{multirow}
\usepackage{makecell}
\usepackage{mathtools}
\usepackage{mathrsfs}
\usepackage{amsmath}
\usepackage{dsfont}
\usepackage{microtype}
\usepackage{nicefrac}       % compact symbols for 1/2, etc.

\usepackage[toc,page]{appendix}

\usepackage{hyperref}

\newcommand\myeq{\stackrel{\mathclap{\normalfont\mbox{i.i.d}}}{=}}

\title{Variational Auto-Decoder: A Method for \\ Neural Generative Modeling from Incomplete Data}

\author{%
  Amir Zadeh\\
  LTI, School of Computer Science\\
  Carnegie Mellon University\\
  Pittsburgh, PA 15213 \\
  \texttt{abagherz@cs.cmu.edu} \\
  \And
  Yao-Chong Lim\\
  School of Computer Science\\
  Carnegie Mellon University\\
  Pittsburgh, PA 15213 \\
  \texttt{yaochonl@cs.cmu.edu} \\
  \And
  Paul Pu Liang\\
  MLD, School of Computer Science\\
  Carnegie Mellon University\\
  Pittsburgh, PA 15213 \\
  \texttt{pliang@cs.cmu.edu} \\
  \And
  Louis-Philippe Morency\\
  LTI, School of Computer Science\\
  Carnegie Mellon University\\
  Pittsburgh, PA 15213 \\
  \texttt{morency@cs.cmu.edu} \\
}

\begin{document}

\maketitle

\begin{abstract}
In many machine learning scenarios, missing data is widespread and only incomplete data is observable. Conditioning posterior inference via neural architectures on such volatile inputs is statistically non-trivial, and can affect the generative likelihood. In this paper, we present a method for posterior approximation without the need for direct conditioning on potentially volatile input. This approach, called Variational Auto-Decoder (VAD), falls within the umbrella of Variational Bayes (VB) methods. We establish VAD as a unique tool for generative modeling and representation learning from incomplete data, which: 1) robustly performs approximate posterior inference across various missing rates, 2) maximizes the Evidence Lower Bound (ELBo) effectively even in extreme missing rates, and 3) is robust to distributional shift in missingness (e.g. sudden test-time change in missing rate). In comparison, VAD shows superior reconstruction likelihood than competitive VAE and GAN baselines, with a widening gap between the models as missing rate increases. 
\end{abstract}
\section{Introduction}

Missing data is widespread in machine learning. In many datasets, values can be missing due to factors such as sensor noise or simply the nature of the problem being studied (e.g. self-occlusions in computer vision). The input volatility caused by missing data is non-trivial. In a Variational Bayes (VB) framework, this input volatility is particularly concerning since without accurate approximate posterior inference, the Evidence Lower Bound (ELBo) may not be maximized efficiently (as studied in this paper). Most common VB inference methods use a parametric class of models for approximate posterior inference. Usually, these approaches require sub-optimally imputing the data at the input level, and subsequently expecting the model to generate the near-optimal output (VAEAC~\cite{ivanov2018variational}). This is also shared by GAN imputation methods through the discriminators taking suboptimal input (GAIN~\cite{yoon2018gain}). The input volatility can in turn overwhelm the parameter-limited inference network or discriminator, and subsequently results in inferior imputation performance. 

In this paper, we propose a new Variational Bayes imputation approach. As opposed to imputation by directly conditioning on a parametric model, we propose that approximate posterior inference be performed using gradient-based approaches to maximize the VB objective. Therefore, the need for model-based approximate posterior inference is eliminated, with only a mapping from the latent space to the input space being essential. The proposed approach is called a Variational Auto-Decoder (VAD). VAD has several appealing properties, which are studied in detail in this paper. Comparisons between VAD, VAE, and GAIN methods are also established across multiple publicly available datasets spanning facial landmark generation, multimodal prediction, visual imputation and synthetic data.

\section{Related Work}\label{sec:related}

Learning from incomplete data is a fundamental research area in machine learning. Notable related works fall into categories as denoted below. 

In a neural framework, Variational Auto-Encoders (VAE) have been commonly used for learning from incomplete data~\cite{mccoy2018variational, williams2018autoencoders, nazabal2018handling}. A particular implementation based on Conditional Variational Auto-Encoders (C-VAE) has shown to achieve superior performance over existing methods for learning from incomplete data \cite{ivanov2018variational}. Generative Adversarial Networks (GANs) have been used for missing data imputation~\cite{yoon2018gain}. In certain cases such methods can be hard to train~\cite{salimans2016improved} and may not perform well when modeling data with high missing rate (studied in this paper). 

Application-specific approaches exist in computer vision for image inpainting~\cite{pathak2016context, yang2017high}. These approaches are effective for images, as they are particularly engineered for visual tasks. Such approaches may require similar train and test-time missing rates for best performance.

Previously proposed Markov-chain based approaches require computationally heavy sampling time and full data to be observable during training~\cite{rezende2014stochastic, sohl2015deep, bordes2017learning}. One appeal of these models is that they can directly maximize the evidence (as opposed to the lower bound). Other approaches have relied on simple, yet elegant, learning techniques such as Gaussian Mixture Models~\cite{delalleau2012efficient}, Support Vector Machines~\cite{pelckmans2005handling} or Principle Component Analysis~\cite{dray2015principal}. Such models have fallen short in the recent years due to lacking the necessary complexity to deal with increasingly non-linear nature of many real-world datasets. 
\section{Model} \label{sec:model}

In this section, we outline a parametric neural VB method which falls close to AEVB algorithm, but specifically for the case of incomplete data. We call this approach Variational 
Auto-Decoder (VAD) since it does not utilize a parametric encoder to infer the paramaters of the approximate posterior. We first outline the problem formulation, and subsequently outline the training and inference procedure for VAD. 

\subsection{Problem Formulation}\label{sec:problemformulation}

We assume a ground-truth random variable $\hat{x} \sim p(\hat{x});\ \hat{x} \in \mathbb{R}^{d}$, sampled from a ground-truth distribution, with $d$ being the dimension of the input space. Unfortunately, the space of $\hat{x}$ is considered to not be fully observable. The part that is observable we denote via random variable $x$, regarded hereon as \textit{incomplete} data. We assume that a random variable $\alpha \sim p(\alpha); \alpha \in \{0,1\}^{d}$ denotes whether or not the data is observable through an indicator in each dimension with value $1$ being observable and $0$ being missing. 

We formalize the process of generating the random variable $x$ as the process of first drawing a ground-truth data sample from $p(\hat{x})$ and a missingness pattern sample from $p(\alpha)$, and subsequently removing information from $\hat{x}$ using $\alpha$. We draw $N$ i.i.d. samples from the above process to build a dataset.\footnote{Notably, each datapoint can have a distinct missingness pattern.} For the rest of this paper, the incomplete dataset is regarded as $X=\{x_1, \dots ,x_N\}$ and the missingness patterns are regarded as $A=\{\alpha_1,\dots,\alpha_N\}$. The ground-truth dataset is regarded as $\hat{X}=\{\hat{x}_1, \dots ,\hat{x}_N\}$. $\hat{X}$ can never directly be a part of training, validation or testing since it is considered strictly unknown. It may only be used for evaluation purpose. 

\subsection{Training}\label{subsec:training}

Assuming that data distribution $p(x)$ can be approximated using a parametric family of distributions with the parameters $\theta$, learning can be done by maximizing the likelihood $p(X ; \theta)$, w.r.t $\theta$. In practice, the log of the likelihood is often calculated and used. In a latent variable-modeling framework the evidence can often be defined by marginalizing a latent variable as follows:
\begin{equation}\label{eq:earlylike}
    \mathcal{L}(\theta | X)\ \ \myeq \ \ { \displaystyle \sum_{i=1}^N \ \textrm{ln} \  p(x_i ; \theta)} = {\displaystyle \sum_{i=1}^N \ \textrm{ln} \int p(z,x_i ; \theta)\ dz}
\end{equation}
In practice calculating the marginal integral over $p(z,x_i;\theta)$ is either expensive or intractable. Subsequently direct latent posterior inference using $p(z|x_i;\theta)$, which is an essential step in latent variable modeling, becomes impractical. 

For any given $x_i$ and any conditional density $q(z|x_i; \phi)$ with $z$ as an unobserved random variable and $\phi$ as parameters of $q$, we can rewrite the evidence in Equation \ref{eq:earlylike} as follows:
\begin{align}\label{eq:big}
\begin{split}
    \mathcal{L}(\theta ; x_i) 
    &=-\int q(z|x_i; \phi)\ \textrm{ln} \ \frac{p(z|x_i;\theta)}{q(z|x_i; \phi)} \ dz \\ &\quad + \int q(z|x_i; \phi)\ \textrm{ln} \ \frac{p(z|x_i;\theta) p(x_i;\theta)}{q(z|x_i; \phi)} \ dz
\end{split}
\end{align}
With the condition that $q(z|x_i, \phi)>0$ if $p(z|x_i,\theta)>0$. To simplify notation, we refer to true posterior $p(z | x_i; \theta)$ as $p_\theta(z|x_i)$ and approximate posterior $q(z|x_i; \phi)$ as $q_\phi(z|x_i)$. More simply, the likelihood in Equation \ref{eq:big} can be written as:
\begin{equation}\label{eq:likelihood}
    \mathcal{L}(\theta | x_i)=\textrm{KL}\Big(q_\phi(z|x_i) \ \Big|\Big| \ p_\theta(z|x_i)\Big) + \mathcal{V} (q_\phi,\theta|x_i)
\end{equation}
In the above equation, $\textrm{KL}(\cdot \left|\right| \cdot)$ is the Kullback-Leibler divergence. One can directly minimize this asymmetric divergence and approximate the true posterior using an approximate posterior $q_\phi(\cdot)$. However, doing so requires samples to be drawn from the true posterior. Markov Chain Monte Carlo (MCMC) approaches can be used to draw samples from the true posterior, however, such approaches are usually very costly. 

$\mathcal{V}(\cdot)$ is referred to as the Evidence Lower Bound (ELBo) or simply variational lower-bound. 

Through the above formulation, rather than employing a method for learning model parameters through likelihood of data, VB methods approximate the posterior probability $p_\theta(z|x_i)$ with a simpler distribution $q_\phi(z|x_i)$. Lower bound $\mathcal{V}(\cdot)$ can be written as:
\begin{equation}\label{eq:elbo}
    \mathcal{V} (q_\phi,\theta|x_i)=\mathbb{E}_{q_\phi(z|x_i)} \Big[\textrm{ln}\ p_\theta(x|z)\Big] - \textrm{KL} \Big(q_\phi(z)\ \Big|\Big|\ p_\theta(z)\Big)
\end{equation}

In the above equation, the first term encourages the latent samples to show high expected likelihood (through reconstruction of $x_i$) under the approximate posterior distribution, and the second term encourages the latent samples to simultaneously follow the latent prior $p_\theta(z)$. 

For a Variational Auto-Decoder framework, the approximate posterior distribution is not parameterized by a neural network, but rather using a well-known distribution in a free-form manner. Therefore, as opposed to randomly initializing the weights of the encoder, we randomly initialize the intrinsic parameters of the approximate posterior. We focus on the family of multivariate Gaussian distributions for this purpose, however other distributions can also be used, as long as the reparameterization trick~\cite{kingma2013auto} can be defined for them (e.g. Beta, Laplace, Uniform). We define a multivariate Gaussian approximate posterior as:
\begin{equation}\label{eq:qnormal}
q_\phi(z|x_i)\coloneqq \mathcal{N}(z; \mu_i,\Sigma_i)
\end{equation}
Note that $\phi=\{\mu_i,\Sigma_i\}$ are learnable parameters of this approximate posterior distribution. The reparameterization of this posterior is essentially defined as $z=\mu_i+\epsilon\cdot \Sigma_i$ with $\epsilon \sim \mathcal{N}(0,I)$. Using this reparameterization, the gradient of the lower bound $\mathcal{V}(\cdot)$ can be directly backpropagated to the mean $\mu_i$ and variance $\Sigma_i$. 

Likelihood is similarly defined as a multivariate Gaussian with missing dimensions of $x_i$ marginalized:
\begin{equation}
\label{eq:invdecoderp}
    p(x|z,\Lambda_i)=\mathcal{N}\left(\mathcal{F}(z;\theta); x_i, \Lambda_i \right)
\end{equation}
This density is centered around $x_i$ as its mean. The covariance $\Lambda_i$ is defined as a diagonal positive semi-definite matrix with $\alpha_i$ on its main diagonal whenever $\alpha_i \neq 0$. $\mathcal{F}(z;\theta)$ is a neural decoder which takes in the samples drawn from posterior in Equation \ref{eq:qnormal}. The optimization subsequently follows steps similar to EM: first sampling from the approximate posterior to calculate a Monte-Carlo estimate of the lower bound $\mathcal{V}(\cdot)$, and subsequently maximizing w.r.t $\theta$ and $\phi$ (Equations \ref{eq:qnormal} and \ref{eq:invdecoderp}). Algorithm \ref{alg:training} summarizes the training (as well as inference in the Section \ref{subsec:inference}). 

\subsection{Inference}\label{subsec:inference}
Typically, once a generative model is learned, it is used to sample data which belong to the underlying learned distribution. Sampling can be done by sampling from the latent space and subsequently using the decoder to generate the data, with no other steps required. 

In certain cases a new data point is given and the goal is to sample the posterior. Calculating the evidence in Equation \ref{eq:earlylike} is still infeasible, even after training is done. Therefore, for the new datapoint, the same lower bound $\mathcal{V}(\cdot)$ needs to be maximized. Using the same process as during training in Equation \ref{eq:elbo}, the parameters of the approximate posterior are initialized randomly and iteratively updated until convergence. Thus, inference is similar to training, except the learned parameters ($\theta^*$) of the decoder are not updated during inference. For a trained model, the speed of inference process can be substantially improved using recent methods such as PE-SVI~\cite{zadeh2019pseudo}, or meta-gradient approaches~\cite{marino2018iterative}. Once $\mathcal{V}(\cdot)$ is maximized, samples of the approximate posterior can be used to generate similar instances as the given datapoint.

\makeatletter
\def\BState{\State\hskip-\ALG@thistlm}
\makeatother

\begin{algorithm}[t!]
\caption{Training (and inference) process for the VAD models with multivariate normal distribution as approximate posterior.\label{alg:training}}
\begin{algorithmic}[1]
\State $\mathcal{F}: \{\theta^{(0)}\} \gets \textrm{Initialization}$ \Comment{Only for training, gets $\theta^*$ during inference}
\State $q: \{\mu_i^{(0)}, \Sigma_i^{(0)}\} \gets \textrm{Initialization}$ 
\Repeat:
\State $[z] \sim q(z ; \mu^{(t)}_i,\Sigma_i^{(t)})$ \Comment{Sampling approximate posterior, Equation \ref{eq:qnormal}}
\State $[p(x|z;\theta^{(t)})]=\mathcal{N}(\mathcal{F}([z],\theta^{(t)}); x_i, \Lambda_i)$ \hfill \Comment{Equation \ref{eq:invdecoderp}}
\State $\{\theta, \mu_i, \Sigma_i\}^{(t+1)} \gets \underset{\theta, \mu_i,\Sigma_i}{\text{grad\_step}}  \Big(\mathcal{V}\big(q^{(t)},\theta^{(t)}|x_i \big)\Big)$ \Comment{No $\text{grad\_step}$ w.r.t $\theta$ during inference}
\State $t \gets t+1$
\Until{maximization convergence on $\mathcal{V}$}
\end{algorithmic}
\end{algorithm}
\section{Experiments}\label{sec:experiments}
\begin{figure*}[t]
\centering{
\includegraphics[width=\linewidth]{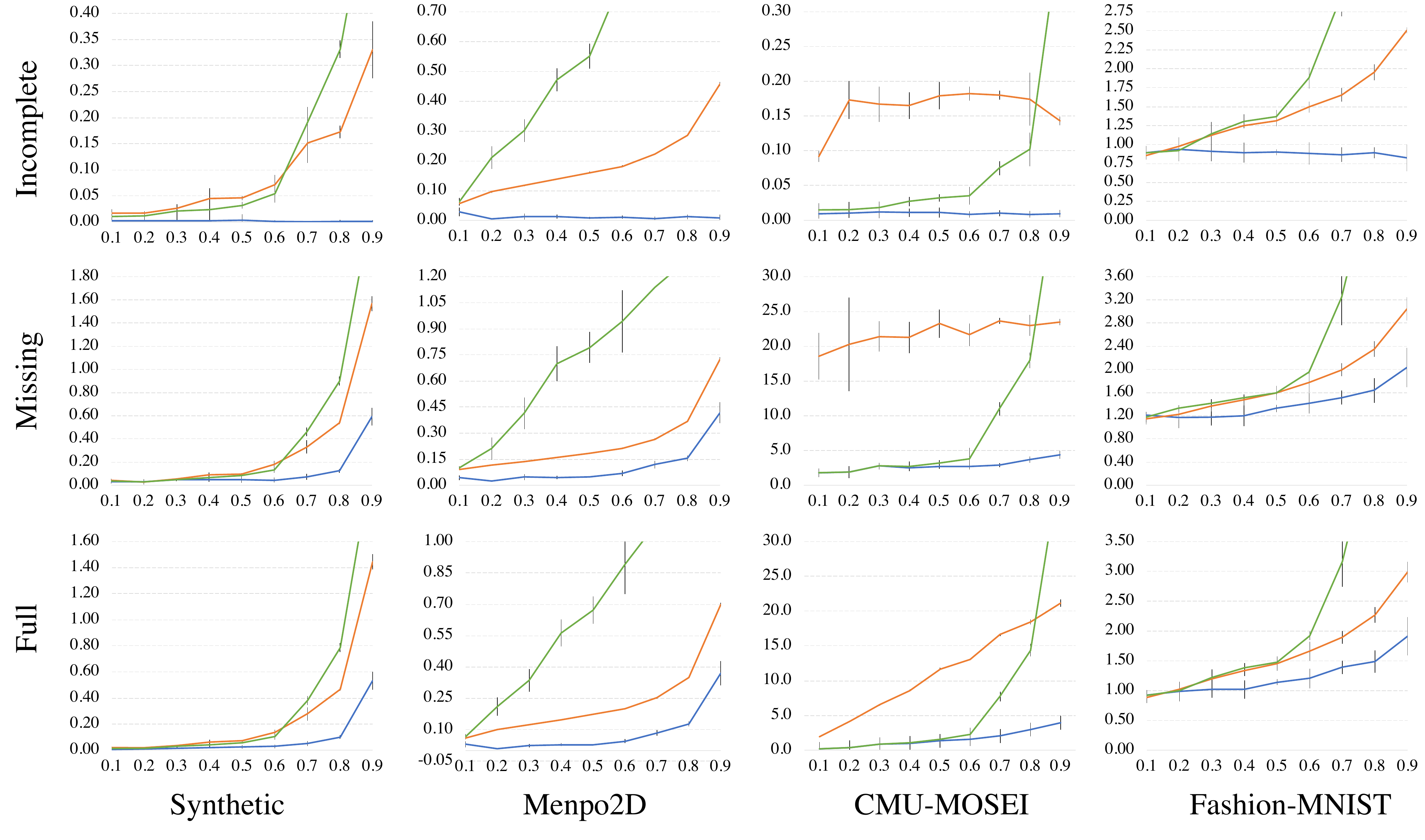}
\caption{\label{fig:results}The results of the Experiment 1 (Section \ref{subsec:x1}) for incomplete, missing and full categories. Best viewed zoomed in and in color. Blue curve shows VAD results, orange curve shows VAE results, and green curve shows the GAIN results. The $x$ axis denotes the missing rate $r$ and the $y$ axis is the reconstruction Mean Squared Error (MSE, lower better). The standard deviation is calculated based on $5$ test runs of the best performing model on the validation set. The gap between both models becomes larger as the missing ratio $r$ increases. The performance thresholds from left to right are: $5.46$, $0.62$, $50.83$, $8.64$ indicating the MSE beyond which models are predicting worse than average of each dimension regardless of input. }}
\end{figure*}
% In the full category, $r=0.0$ shows the performance of the case where there is no missing data (train or test).
Based on the Equation \ref{eq:elbo}, the variational lower bound $\mathcal{V}(\cdot)$ is dependent on the expectation of the log likelihood $p(x|z)$ under the approximate posterior $q_\phi(z|x_i)$. This term, which relies on the incomplete input $x_i$, indicates how well samples drawn from the approximate posterior are able to recreate $x_i$ (using likelihood in Equation \ref{eq:invdecoderp}). If due to the parameterization of the approximate posterior, this term cannot be maximized efficiently for incomplete data during training, then maximizing the lower bound will subsequently be impacted.\footnote{In simple terms, regardless of the second term in Equation \ref{eq:elbo}, if the approximate posterior and decoder cannot reproduce the data efficiently in the best case, then generative modeling will not be successful, regardless of the second term of Equation \ref{eq:elbo} (which is the same for both models).} For both VAD and VAE, we aim to address whether missing data can cause issues for maximizing this expectation. Therefore, we specifically study the lower bound $\mathcal{V}(\cdot)$ with only the first term to compare if any of the two models inherently fall short in presence of missing data.

In the experiments,\footnote{Code and data available through \url{https://github.com/A2Zadeh/Variational-Autodecoder}} VAD, VAE and GAIN are trained on identical data. VAE closely follows the Arbitrary Conditioning method~\cite{ivanov2018variational}; a state-of-the-art VAE training process specifically designed for imputation. VAD and VAE maximize the same lower bound $\mathcal{V}(\cdot)$. The only difference between the two models is therefore parameterization of the approximate posterior distribution: for VAD it is the parameters of the distributions and for VAE it is the weights of the neural encoder. A validation set is used to choose the best performing hyperparameter\footnote{Both models undergo substantial hyperparameter search as described in Appendix \ref{app:imp} (with exact values).} setup exactly based on the lower bound $\mathcal{V}(\cdot)$. Subsequently, the best trained model is used on test data. The ground truth is never used during training, validation or testing (unless required by the experiment, described later in this section). Only for evaluation purposes, after the inference is done on test set, the ground truth is simply \textit{revealed}. To report a measure that is easy to compare, we report the MSE\footnote{MSE is calculated per each dimension, therefore it is independent of the missing rate.} (Mean Squared Error) between the decoded mean of the approximate posterior in the following categories: 1) Incomplete: we report the MSE between the incomplete data (available dimensions) and the output of the decoder. Since the incomplete data is the basis of the likelihood in Equation \ref{eq:invdecoderp}, we expect models to show low MSE for the incomplete data. 2) Missing: once the inference is done over the incomplete data, missing values are revealed to evaluate the imputation performance of both models. 3) Full: after revealing missing values, we can simply calculate the performance over the full ground-truth data.

Specifically, the following two scenarios are studied in the form of experiments in this paper:

\textbf{Experiment 1:} We study the case where during train and test time, data follows a similar missing rate. Essentially the distribution of missingness is also the same for these cases. 

\textbf{Experiment 2:} In real-world situations it is very unlikely that the data will follow the same missing rate during train and test time. Therefore, we study 3 cases: a) only test-time missingness, which shows how robust the learned models are in deployment. b) only train-time missingness, which shows if models can effectively learn the underlying true distribution even when given incomplete data during training, and b) train and test time both missing but with different rates, which shows combinations of properties of (a) and (b).

\subsection{Datasets}

We experiment with a variety of datasets from different areas within machine learning. To better understand the ranges of MSE for each dataset, we report a baseline obtained by taking the mean of the ground-truth training data as the prediction during test time. This threshold indicates the limit beyond which models are performing worse than just projecting the mean of the ground-truth data regardless of the input. 

\begin{figure*}[t]
\centering{
\includegraphics[width=\linewidth]{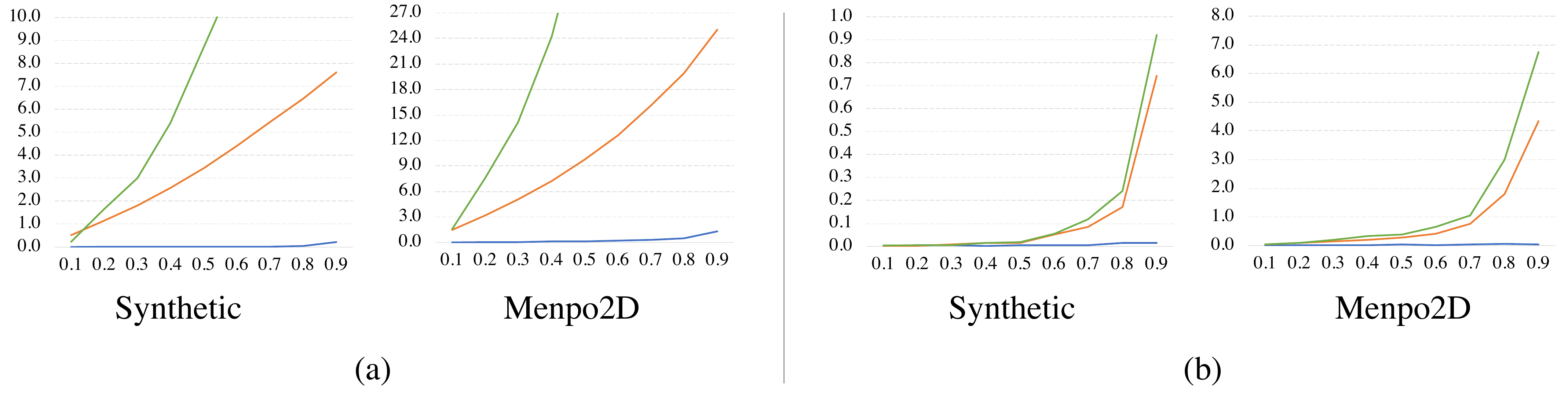}
\caption{\label{fig:ex2}Best viewed zoomed in and in color. Results (in Full category) of Experiment 2 (Section \ref{subsec:x2}) for (a) test-time missingness and (b) train-time missingness. Blue curve shows VAD results, orange curve shows VAE, and green curve shows GAIN results. The $x$ axis denotes the missing rate $r$ and the $y$ axis is the reconstruction Mean Squared Error (MSE). In both the experiments VAD shows superior performance than VAE. The performance of VAE is significantly affected in both scenarios. }}
\end{figure*}
\textbf{Toy Synthetic Dataset:} We study a case of synthetic data where we control the distributional properties of the data. In the generation process, we first acquire a set of independent dimensions randomly sampled from 5 univariate distributions with uniform random parameters: \{\texttt{Normal, Uniform, Beta, Logistic, Gumbel}\}. Often in realistic scenarios there are inter-dependencies among the dimensions. Hence we proceed to generate interdependent dimensions by picking random subsets of the independent components and combining them using random operations such as weighted multiplication, affine addition, and activation. Using this method, we generate a dataset containing $N=50,000$ datapoints with ground-truth dimension $d=300$. Further details of the generation and exact ranges are given in the supplementary material. The threshold MSE for this dataset is $5.46$.

\textbf{Menpo Facial Landmark Dataset:} Menpo2D contains $13,391$ facial images with various subjects, expressions, and poses~\cite{zafeiriou2017menpo}. Due to these variations, the nature of the dataset is complex. Since Menpo dataset has ground truth annotations for $84$ landmarks regardless of self-occlusions in the natural image, it allows for creating a real-life ground-truth data for our experiments. The purpose of using this dataset is to compare the two models on how well they recreate the structure of an object given only a subset of available keypoints. The threshold MSE for this dataset is $0.62$.

\textbf{CMU-MOSEI Dataset:} CMU Multimodal Sentiment and Emotion Intensity (CMU-MOSEI) is an in-the-wild dataset of multimodal sentiment and emotion recognition \cite{zadeh2018multimodal}. The datasets consist of sentences utterance from online YouTube monologue videos. CMU-MOSEI consists of 23,500 such sentences and with three modalities of text (words), vision (gestures) and acoustic (sound). For text modality, the datasets contains GloVe word embeddings~\cite{pennington2014glove}. For visual modality, the datasets contains facial action units, facial landmarks, and head pose information. For acoustic modality, the datasets contain high and low-level descriptors following COVAREP~\cite{degottex2014covarep}. We use expected multimodal context for each sentence, similar to unordered compositional approaches in NLP~\cite{iyyer2015deep}. The threshold MSE for this dataset is $50.83$.

\textbf{Fashion-MNIST:} Fashion-MNIST\footnote{\url{https://github.com/zalandoresearch/fashion-mnist}} is a variant of the MNIST dataset. It is considered to be more challenging than MNIST since variations within fashion items are usually more complex than written digits. The dataset consists of $70,000$ grayscale images with shape $28\times28$ from 10 fashion items. The threshold MSE is $8.64$ for this dataset. 

We base our experiments on Missing Completely at Random (MCAR), which is a severe case of missingness. For each $\hat{x_i}$, we sample a missing pattern $\alpha_i \sim \texttt{Bernouli}(r); \alpha_i \in \{0,1\}^d$ with missing ratio $r$ ranging from $0.1$ to $0.9$ with increments of $0.1$. This form of $\alpha_i$ essentially allows each dimension to unexpectedly go missing. 

\subsection{Experiment 1: Similar Train and Test Missing Rate}\label{subsec:x1}

In this experiment, both train and test data follow the same missing ratio $r$. For each $r$, models are trained using likelihood in Equation \ref{eq:invdecoderp} and maximize the lower-bound $\mathcal{V}(\cdot)$ in Equation \ref{eq:elbo}. Figure \ref{fig:results} shows the results of this experiment for the best validated models for both VAD and VAE. Overall, in all the three incomplete, missing and full categories, VAD shows superior performance than VAE and GAIN. As the missingness increases, the gap between the two models widens in all three categories (except for CMU-MOSEI where the performance gap is large in incomplete and missing even for small $r$). This essentially demonstrates that VAE and GAIN become increasingly unstable in presence of missing data. Specifically for the case of incomplete data, VAE is not able to perform reliable posterior inference since the output of the decoder increasingly deviates farther away from the available input. VAD on the other hand, shows steady performance in the incomplete category. The performance of all the models are naturally affected for the missing category as $r$ increases (in reality some of the missing data may not be imputable given the available input). However, the increasing gap between the two models also appears in missing category. Finally the comparison in the full category shows that VAD is able to regenerate the ground-truth better than VAE. In a full picture, Figure \ref{fig:results} suggests that approximate posterior using an encoder conditioned on a volatile input becomes increasingly unstable as missingness becomes more severe.

\begin{figure*}[t]
\centering{
\includegraphics[width=.9\linewidth]{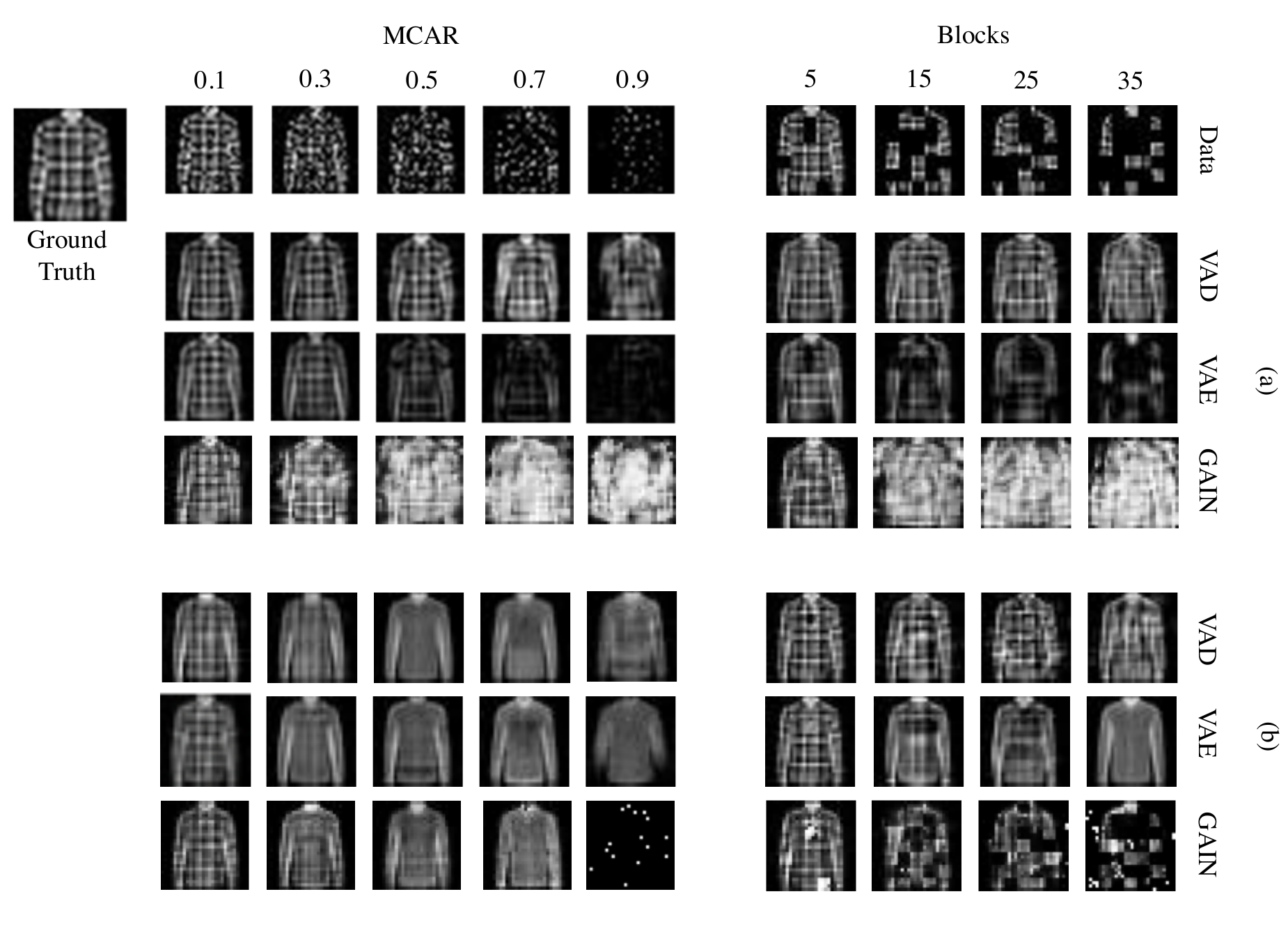}
\caption{\label{fig:inapint}Visualization of  Fashion-MNIST inpainting. Top row (Data) shows the given data to VAD, VAE, and GAIN. Ground-truth is the real image from Fashion-MNIST. For (a), training is done on ground-truth data and testing is done on incomplete data. For (b) training and testing is done on similar missing rate. For the case of MCAR (a), performance of VAE and GAIN significantly deteriorate when trained on ground-truth data and tested on incomplete data. This trend is also visible but at a much slower rate for the case (b) - where only at $r=0.9$ VAE and GAIN shows visually perceivable failure. We also compare the performance of both models when the missingness changes from MCAR random missing $4\times4$ blocks - training for case (a) still done on ground-truth and (b) done in presence of random blocks. }}
\end{figure*}  
\begin{figure*}[t]
\centering{
\includegraphics[width=.9\linewidth]{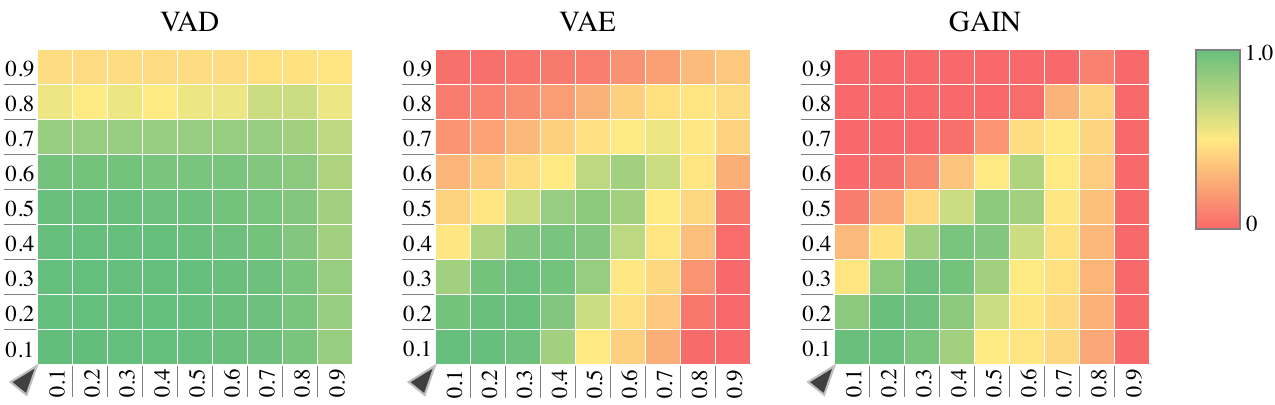}
\caption{\label{fig:heatmap}Sensitivity experiments for training and validation on varying missing rates, for Fashion-MNIST dataset. $x$ dimension is the validation missing rate, and $y$ dimension is the training missing rate. Sensitivity is expressed in terms of likelihood of the reconstructed samples in validation set (inference is done on partial data, and likelihood calculation is done on full data). VAD remains robust.}}
\end{figure*}  

\subsection{Experiment 2: Different Train and Test Missing Rate}\label{subsec:x2}

While in the previous experiment both the train and test stages followed the same missing rate, realistic scenarios are often more complex. 

\textbf{Test-time missingness:} In the test-time missingness scenario, models are trained on the complete data. During testing, the data goes missing following a missing ratio $r$ ranging from $0.1$ to $0.9$. Figure \ref{fig:ex2} shows the results of this experiment for the synthetic and Menpo2D datasets in Full category. In both dataset the performance of VAE and GAIN is substantially affected by the missing data during testing, specially when compared to Experiment 1 (trained and tested on the same missingness). The performance of VAD remains almost similar for both synthetic and Menpo2D datasets and relatively similar to Experiment 1. We also visually demonstrate this in an inpainting scenario. We compare models when they are trained on ground-truth and tested on incomplete data against when they are trained on similar missing ratio. Both models are trained on the Fashion-MNIST ground-truth train set and subsequently during testing the data may go missing. Figure \ref{fig:inapint} shows the testing performance of the models for different missing ratios as well as block-sized missingness. Visually, it can be seen that VAD shows superior performance compared to baselines. Compared to the case where missing ratio is the same, we observe that it is crucial for VAE and GAIN to train on the same missing rate as the test time. 

\textbf{Train-time missingness:} The train-time missingness scenario is the opposite of the above scenario\footnote{Not to be confused with denoising methods or dropout which map noisy input to the ground-truth during train time. In this scenario the ground-truth training set can never be fully observed for training.}. Models are trained on a train set with a missing ratio $r$ ranging from $0.1$ to $0.9$. During testing, they perform inference on a different missing rate, in the extreme case on ground-truth test. The right side of Figure \ref{fig:ex2} shows the results of this experiment on VAD, VAE and GAIN. We observe a similar trend of performance between the models, with VAD remaining consistent while the performance of baselines deteriorates as missing rate increases. 

\textbf{Sensitivity analysis:} Figure \ref{fig:heatmap} demonstrates the sensitivity of each model to the train and test missing rate mismatch, for Fashion-MNIST dataset. One key observation is that both VAE and GAIN only show high likelihood of the data around the diagonal where train and test missing rates are similar. This trend is not present in VAD, which shows high likelihood across different rates.

\section{Conclusion}

In this paper, we presented a Variational Bayes method for the case of incomplete data, called Variational Auto-Decoder (VAD). We studied the effect of missing data on the approximate posterior conditioned directly using an encoder on the incomplete input. We showed that such conditioning (VAE) may not allow for maximizing the lower bound efficiently, due to poor performance for maximizing the expected likelihood (under the approximate posterior) of the incomplete data. Similar effect was observed on GAIN, where the discriminator receives volatile input. The approximate posterior in VAD is parameterized by a known distribution, parameters of which are directly optimized in a VB learning framework. The parameters of the approximate posterior can be learned using gradient-based approaches. We showed that VAD framework has unique and appealing properties for missing data imputation and approximate posterior inference. 

\clearpage
%\input{NeuRIPS2019/sections/impact}

%\clearpage

\bibliographystyle{plain}
\bibliography{neurips.bib}

\clearpage
\begin{appendices}

\section{Implementation Remarks} \label{app:imp}
Here we detail the hyperparameter space used for the experiments. We establish a fair comparison between the VAD, VAE, and GAIN models by using similar hyperparameters. All models underwent substantial hyperparameter space search (see Table \ref{table:hyperparameters}). Validation set of the studied datasets are used for choosing the best models. We varied three main hyperparameters: the dimensionality of the posterior (or noise) space $d_z$, the number of feedforward hidden layers in the decoder $\mathcal{F}$ and the encoder/discriminator (encoder only for VAE), and the number of hidden units in each hidden layer. A summary is shown in Table \ref{table:hyperparameters}. 

During inference of VAD models, we simply stopped once the model reached a plateau. Since VAD models have a high degree of freedom for approximate posterior $q
_\phi(\cdot)$, we observed that it is crucial to use Adam \cite{kingma2014adam} for learning the parameters of the approximate distribution. For all models, learning rates of $10^{\{-2,-3\}}$ for the latent variables and hidden units were used. 

During inference, VAD and VAE models use an approximate posterior with a learnable variance. However, in practice when using incomplete data, learning the variance was a very unstable process for VAE. VAE models showed very high sensitivity to even small variances, quickly performing similar to projecting the mean in each dimension. We believe combination of noise from imputed values and also the noise added through approximate posterior may cause substantial uncertainties for VAE. While VAD models suffered similarly from the same instability during learning the variance, the performance was better than VAE. 

Therefore, during our experiments we treated the approximate posterior variance as a hyperparameter and trained the models for different variances. This way results improved substantially. The best performing variance may change depending on the problem and the range of the input space, however, in general we observed that very large variances did not converge well, while very small variances did not yield the best results.

\subsection{Synthetic Data Generation}
The parameters of the synthetic data are outlined in Table \ref{table:syn_gen_params}.

\newcolumntype{K}[1]{>{\centering\arraybackslash}p{#1}}
\newcolumntype{L}[1]{>{\arraybackslash}p{#1}}
\definecolor{gg}{RGB}{45,190,45}

\begin{table*}[tb]
\fontsize{9}{10}\selectfont
\centering
\begin{tabular}{|L{3.4cm} | L{1cm} | L{1.8cm}|}
    \hline
    Hyperparameter & Group & Values \\
    \hline
    \multirow{2}{*}{\# of latent variables} & A & 25, 50, 100 \\
    & B & 50, 100, 400 \\
    \hline
    \multirow{2}{*}{\# of hidden units per layer} & A & 50, 100, 200 \\
    & B & 100, 200, 400 \\
    \hline
    \# of hidden layers & A, B & 2, 4, 6 \\
    \hline
    LR of network parameters and latent variables & A, B & $10^{-2}, 10^{-3}$ \\
    \hline
\end{tabular}
\caption{Hyperparameters used for the experiments on VAD, VAE and GAIN across different datasets. Group B refers to Fashion-MNIST, while group A refers to all other datasets.}
\label{table:hyperparameters}
\end{table*}

\newcolumntype{K}[1]{>{\centering\arraybackslash}p{#1}}
\newcolumntype{L}[1]{>{\arraybackslash}p{#1}}
\definecolor{gg}{RGB}{45,190,45}

\begin{table*}[tb]
\fontsize{9}{10}\selectfont
\centering
\begin{tabular}{|L{2.2cm} | L{1.2cm} | L{2.2cm}|}
    \hline
    Distribution & Parameter & Range of values \\
    \hline
    \multirow{2}{*}{\texttt{Normal}$(\mu, \sigma)$} & $\mu$ & $[-1, 1]$ \\
    & $\sigma$ & $(0, 2]$ \\
    \hline
    \multirow{2}{*}{\texttt{Uniform}$(a, b)$} & $a$ & $[-2, 2]$ \\
    & $b$ & $[a, 2]$ \\
    \hline
    \multirow{2}{*}{\texttt{Beta}$(\alpha, \beta)$} & $\alpha$ & $[0, 3]$ \\
    & $\beta$ & $[0, 3]$ \\
    \hline
    \multirow{2}{*}{\texttt{Logistic}$(\mu, s)$} & $\mu$ & $[-1, 1]$ \\
    & $s$ & $(0, 2]$ \\
    \hline
    \multirow{2}{*}{\texttt{Gumbel}$(\mu, \beta)$} & $\mu$ & $[-1, 1]$ \\
    & $\beta$ & $(0, 2]$ \\
    \hline
\end{tabular}
\caption{Parameters used during generation of synthetic data.}
\label{table:syn_gen_params}
\end{table*}

\end{appendices}

\end{document}